\documentclass[
twocolumn,
]{ceurart}

\usepackage{listings}
\usepackage{tipa}
\usepackage{soul}
\usepackage[dvipsnames,table]{xcolor}    
\usepackage{booktabs}         
\usepackage{multirow}  
\usepackage{siunitx} 
\usepackage{fancybox}
\usepackage{graphicx}
\usepackage{tabularx}
\usepackage{makecell}
\usepackage{arydshln}
\usepackage{pifont}
\usepackage{hyperref}

\newcommand{\ap}[1]{{\textcolor{black}{#1}}}

\begin{document}

\copyrightyear{2025}
\copyrightclause{Copyright for this paper by its authors.
  Use permitted under Creative Commons License Attribution 4.0
  International (CC BY 4.0).}

\conference{CLiC-it 2025: Eleventh Italian Conference on Computational Linguistics, September 24 — 26, 2025, Cagliari, Italy}

\title{Gender-Neutral Rewriting in Italian: Models, Approaches, and Trade-offs}

\author[1,2]{Andrea Piergentili}[%
orcid=0000-0003-2117-1338,
email=apiergentili@fbk.eu,
]
\cormark[1]
\address[1]{University of Trento, via Sommarive 5, 38123, Povo (TN), Italy}
\address[2]{Fondazione Bruno Kessler, Via Sommarive 18, 38123, Povo (TN), Italy}

\author[2]{Beatrice Savoldi}[%
orcid=0000-0002-3061-8317,
email=bsavoldi@fbk.eu,
]

\author[2]{Matteo Negri}[%
orcid=0000-0002-8811-4330,
email=negri@fbk.eu,
]

\author[2]{Luisa Bentivogli}[%
orcid=0000-0001-7480-2231,
email=bentivo@fbk.eu,
]

\cortext[1]{Corresponding author.}

\begin{abstract}
  Gender-neutral rewriting (GNR) aims to reformulate text to eliminate unnecessary gender specifications while preserving meaning, a particularly challenging task in grammatical-gender languages like Italian. 
  In this work, we conduct the first systematic evaluation of state-of-the-art large language models (LLMs) for Italian GNR, introducing a two-dimensional framework that measures both neutrality and semantic fidelity to the input. We compare few-shot prompting across multiple LLMs, fine-tune selected models, 
  and apply targeted cleaning to boost task relevance. Our findings show that 
  open-weight
  LLMs outperform the only existing model dedicated to GNR in Italian,
  whereas our fine-tuned models match or exceed the best 
  open-weight
  LLM's performance at a fraction of its size. Finally, we discuss the trade-off between optimizing the training data for neutrality and meaning preservation.
\end{abstract}


\maketitle

\section{Introduction}

Language technologies reinforce existing gender stereotypes and binary assumptions by disproportionately favoring masculine references or representations
\cite{savoldi-etal-2024-harm}, 
especially when gender information is ambiguous or unspecified \cite{kotek-2023-genderbias-llms,ostrow2025llmsreproducestereotypessexual,SAVOLDI2025101257}. Such biases result in the under-representation or misrepresentation of certain gender groups, reinforcing existing societal stereotypes, and erasing non-binary identities \cite{blodgett-etal-2020-language,dev-etal-2021-harms}. Addressing these biases through gender-inclusive approaches is increasingly important to ensure language technologies contribute to more inclusive and equitable communication \cite{gabriel-20218-neutralization,apa2020publication,piergentili-etal-2023-gender}.

Gender-neutral rewriting (GNR) has emerged as a natural language generation task aimed at producing texts free from unnecessary gender specifications \cite{sun2021theythemtheirs,vanmassenhove-etal-2021-neutral}. This task is particularly challenging in grammatical-gender languages, such as Italian, due to the pervasive encoding of gender in the morphology. 
Consider the sentence \textit{`Tutt\underline{i} \underline{i} sena\underline{tori} sono stat\underline{i} informat\underline{i}'} (equivalent to \textit{All\textsubscript{M} the\textsubscript{M} senators\textsubscript{M} have been\textsubscript{M} informed\textsubscript{M}}): almost every word is morphologically inflected for (masculine) gender. Rephrasing this sentence in a gender-neutral way 
may require significant changes, e.g. `\textit{Ogni membro del Senato ha ricevuto l'informazione}' (\textit{Every member of the Senate has received the information}).
A further challenge in automatic GNR is preserving the meaning of the original sentence beyond gender expression, to avoid generating output sentences that are neutral but semantically divergent from the input.

So far, GNR system development has been mostly confined to English \cite[inter alia]{sun2021theythemtheirs,vanmassenhove-etal-2021-neutral,bartl-leavy-2024-showgirls}, where gender is expressed through specific sets of words, such as pronouns (e.g., \textit{he/she}, \textit{him/her}) and lexically gendered terms (e.g., \textit{policeman/policewoman}), and gender-neutral alternatives (e.g., the singular \textit{they} or synonyms like \textit{police officer}) are generally available and attested. GNR systems for grammatical-gender languages generally target specific gendered phenomena, such as member nouns \cite{doyen2025genrefrenchgenderneutralrewriting}, or use neologistic \cite{rose_variation_2023} inclusive devices such as neomorphemes and graphemic solutions \cite{pomerenke2022inclusify,veloso-etal-2023-rewriting,lerner-grouin-2024-inclure} that convey neutrality, but are not necessarily acceptable in all contexts.
Currently, the sole model dedicated to Italian GNR was developed by \citet{greco-2025-italian-rewriting}, 
which, however, was developed and tested on proprietary, not publicly available data, hindering reproducibility and progress. 

Towards addressing this gap, this paper explores the potential of state-of-the-art (SOTA) large language models (LLMs) to perform GNR in Italian.
Specifically, we explore both prompting and fine-tuning approaches and assess both neutrality and meaning preservation in the reformulated texts. 

Our contributions are threefold: \textit{\textbf{i)}} The first systematic evaluation of SOTA LLMs for Italian GNR under a two-dimensional framework measuring both neutrality and meaning preservation;
\textit{\textbf{ii)}} A set of experiments in fine-tuning LLMs for GNR,
enabling compact models to rival significantly larger-sized models; 
\textbf{\textit{iii)}} An investigation of the GNR performance trade-off between meaning preservation and neutrality in the outputs of LLMs fine-tuned on sentence similarity-optimized data.\footnote{We release models and data at \url{https://huggingface.co/FBK-MT}}

\section{Background}
\label{sec:background}

\paragraph{Gender-Inclusive Language}

Inclusive language aims to prevent expressions that reinforce gender hierarchies or render non-binary identities invisible, promoting fairness and inclusion in alignment with UN Sustainable Development Goals of gender equality.\footnote{See \url{https://sdgs.un.org/goals/goal5}}
In grammatical-gender languages like Italian, inclusive language is both particularly challenging and increasingly urgent due to their entrenched gender systems \citep{papadopoulos2019morphological, ScottoDiCarlo2024, SilvaSoares2024} and the widespread use of masculine forms as default to mark generic or mixed-gender referents \citep{GYGAX2021101328}.\footnote{English presents fewer challenges as gender marking is primarily limited to pronouns, allowing focused solutions like the singular \textit{they} \citep{ackerman2019syntactic}.}
To address this issue, two main strategies have emerged, as reviewed by \citet{rosola-etal-2023-beyond} within the Italian linguistic context. On the one hand, \textit{innovative} forms using neomorphemes and symbols (e.g., tutt* or tutt@) are mostly used in informal contexts like social media and online LGBTQIA+ communities, and are generally not accepted in more formal contexts \cite{Comandini_2021}. Instead, \textit{conservative} gender-neutral language strategies retool existing forms and grammar to avoid unnecessary gendered expressions \citep{silveira1980generic, bailey2022based}, e.g. by replacing \textit{i professori} with \textit{la docenza} \citep{piergentili-etal-2023-gender}. As attested by \citet{piergentili-etal-2023-hi}, such neutral solutions are increasingly accepted in communication and are endorsed by institutions and universities to embrace all gender identities \citep{hoglund2023gendering}.\footnote{See for instance the EU Parliament guidelines for gender-neutral language: \url{https://www.europarl.europa.eu/cmsdata/151780/GNL_Guidelines_EN.pdf}}

\paragraph{Gender-Inclusive Rewriting}

In recent years, sexism and gender-exclusionary practices have been increasingly addressed in NLP, focusing initially on binary gender bias and more recently expanding to non-binary inclusive language technologies  \citep{dev-etal-2021-harms, SAVOLDI2025101257}. NLP work has explored the modeling of inclusive language across various tasks \citep{cao-daume-iii-2020-toward, waldis-etal-2024-lou}, including inclusive language generation. For instance, \citet{bartl-leavy-2024-showgirls} explored stereotype reduction in English LLMs fine-tuned on inclusive seeds and lexicon.

Intralingual inclusive rewriting has primarily been explored in English \citep{sun2021theythemtheirs, vanmassenhove-etal-2021-neutral, bartl-leavy-2024-showgirls}, where gender marking is scarce. Similar efforts in languages with grammatical gender include research on German \citep{pomerenke2022inclusify},
Portuguese \citep{veloso-etal-2023-rewriting}, and French \citep{lerner-grouin-2024-inclure, doyen2025genrefrenchgenderneutralrewriting}, either by using \textit{innovative} forms or targeting specific instances of gendered languages---such as masculine generics in member nouns.
In Italian, prior work has explored gender-neutral translation \citep{savoldi-etal-2024-prompt, savoldi2025mgente}, whereas intra-lingual rewriting remains mostly limited to benchmarking efforts.
\citep{frenda-etal-2024-gfg}. \citet{attanasio-etal-2024-itaeval} compared several instruction-following models prompted across fairness-related tasks---including GNR---but these underperformed, achieving less than 50\% success in neutralization. 
\citet{frenda-etal-2024-gfg} proposed the gender-fair generation (GFG) challenge,
where for one of the tasks models are prompted to reformulate gendered Italian sentences in a neutral way.
Closest to our work, \citet{greco-2025-italian-rewriting} developed a rewriter by fine-tuning language models specifically for Italian gender-neutral language.
However, the data used for testing and developing these models are not publicly available, hampering further research and comparability.

\begin{table}[t]
\footnotesize
\centering
\begin{tabular}{cl}
\toprule
\textbf{REF-G} & Spero di essere \underline{stato chiaro} su questo punto.
\\
EN & I hope that I am clear in this. \\
\midrule
\textbf{REF-N} & Spero di \textit{avere espresso con chiarezza} questo punto.
\\
EN & I hope that I have expressed this point clearly. \\
\bottomrule
\end{tabular}
\caption{Example of an Italian \textsc{mGeNTE} entry. The gendered words in the REF-G are underlined, the corresponding neutralization In REF-N is italicized.}
\label{tab:mgente-examples}
\end{table}

\section{Experimental settings}
\label{sec:settings}

We define GNR as the task of reformulating a sentence to remove explicit gender markings referring to human entities, without altering the sentence beyond what is necessary for neutralization, ensuring semantic equivalence to the input.
We run a set of experiments evaluating different systems and approaches to GNR. Here, we first
discuss the evaluation data and metrics (§\ref{sec:settings-eval}) and the set of models we experiment with (§\ref{sec:settings-models}). Then, we describe two approaches to GNR: few-shot prompting SOTA LLMs (§\ref{sec:settings-prompting}) and fine-tuning a subset of those LLMs on repurposed Italian data (§\ref{sec:settings-finetuning}).

\setlength{\tabcolsep}{4pt}
\begin{table*}[ht]
\footnotesize
\centering
\begin{tabular}{llcccccc}
\toprule
  \textbf{Group} 
  & \textbf{Model} 
  & \textbf{Size (B)} 
  & \textbf{Prompting} 
  & \textbf{Fine-tuning} 
  & \textbf{Paper / Report}
  & \textbf{Link} \\
\midrule

\multirow{3}{*}{\textbf{"Italian" models}} 
  & Minerva     &  7  & \color{ForestGreen}{\ding{52}} & \color{WildStrawberry}{\ding{56}} & \citet{orlando-etal-2024-minerva} & \href{https://huggingface.co/sapienzanlp/Minerva-7B-instruct-v1.0}{\includegraphics[height=1em]{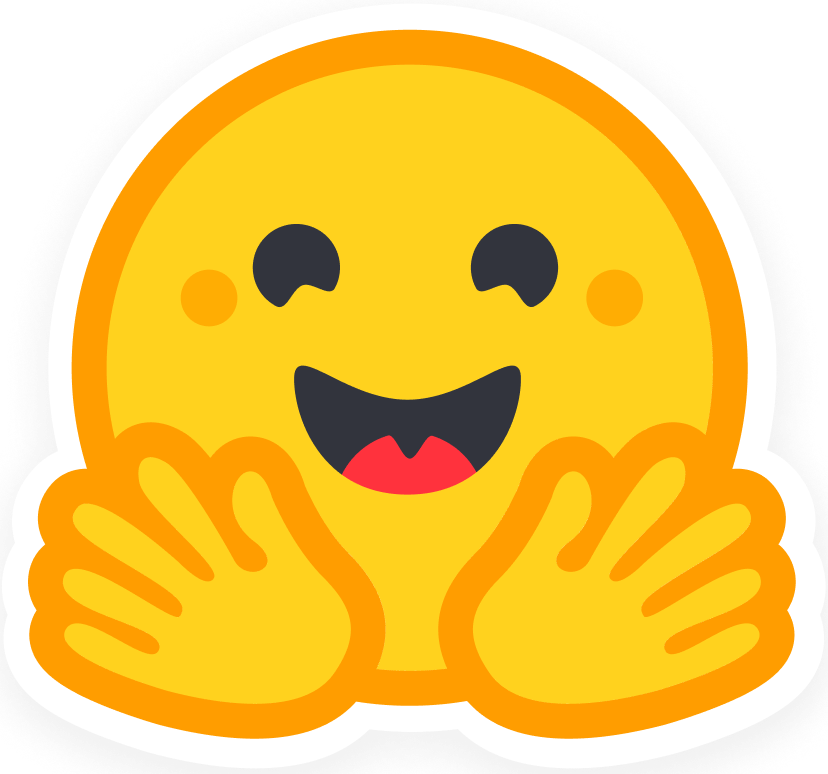}}\\
  & LLaMAntino  &  8  & \color{ForestGreen}{\ding{52}} & \color{ForestGreen}{\ding{52}} & \citet{basile2023llamantino} & \href{https://huggingface.co/swap-uniba/LLaMAntino-3-ANITA-8B-Inst-DPO-ITA}{\includegraphics[height=1em]{figures/hf-logo.png}} \\
  & Velvet      & 14  & \color{ForestGreen}{\ding{52}} & \color{ForestGreen}{\ding{52}} & \citet{almawave_velvet_2025} & \href{https://huggingface.co/Almawave/Velvet-14B}{\includegraphics[height=1em]{figures/hf-logo.png}} \\
\midrule

\multirow{3}{*}{\textbf{Multilingual LLMs}} 
  & Llama 3.1   &  8  & \color{ForestGreen}{\ding{52}} & \color{ForestGreen}{\ding{52}} & \citet{meta2024llama3herdmodels}  & \href{https://huggingface.co/meta-llama/Llama-3.1-8B-Instruct}{\includegraphics[height=1em]{figures/hf-logo.png}} \\ 
  & Phi 4       & 14  & \color{ForestGreen}{\ding{52}} & \color{ForestGreen}{\ding{52}} & \citet{abdin2024phi4technicalreport} & \href{https://huggingface.co/microsoft/phi-4}{\includegraphics[height=1em]{figures/hf-logo.png}} \\
  & Llama 3.3   & 70  & \color{ForestGreen}{\ding{52}} & \color{WildStrawberry}{\ding{56}} & \citet{meta2024llama3herdmodels}  & \href{https://huggingface.co/}{\includegraphics[height=1em]{figures/hf-logo.png}} \\
\hdashline

\multirow{4}{*}{\textbf{Qwen3 family}} 
  & Qwen3 &  4  & \color{ForestGreen}{\ding{52}} & \color{WildStrawberry}{\ding{56}} & \multirow{4}{*}{\citet{qwen3technicalreport}} & \href{https://huggingface.co/Qwen/Qwen3-4B}{\includegraphics[height=1em]{figures/hf-logo.png}} \\
  & Qwen3 &  8  & \color{ForestGreen}{\ding{52}} & \color{ForestGreen}{\ding{52}} &  & \href{https://huggingface.co/Qwen/Qwen3-8B}{\includegraphics[height=1em]{figures/hf-logo.png}} \\
  & Qwen3 & 14  & \color{ForestGreen}{\ding{52}} & \color{ForestGreen}{\ding{52}} &  & \href{https://huggingface.co/Qwen/Qwen3-14B}{\includegraphics[height=1em]{figures/hf-logo.png}} \\
  & Qwen3 & 32  & \color{ForestGreen}{\ding{52}} & \color{WildStrawberry}{\ding{56}} &  & \href{https://huggingface.co/Qwen/Qwen3-32B}{\includegraphics[height=1em]{figures/hf-logo.png}} \\
\midrule

\multirow{1}{*}{\textbf{Commercial system}} 
  & GPT 4.1   
    & \multicolumn{1}{c}{?}
    &  \color{ForestGreen}{\ding{52}} & \color{WildStrawberry}{\ding{56}} & \citet{openai2025gpt41} & - \\
\midrule

\multirow{1}{*}{\textbf{Dedicated model}} 
  & Inclusively 
    &  0.78 & \ding{83} & \color{WildStrawberry}{\ding{56}} & \citet{greco-2025-italian-rewriting} & \href{https://huggingface.co/E-MIMIC/inclusively-reformulation-it5}{\includegraphics[height=1em]{figures/hf-logo.png}} \\
\bottomrule
\end{tabular}
\caption{Summary of the models used in this work, including their size, usage in prompting and fine-tuning experiments, and documentation. Inclusively (\ding{83}) is a sequence-to-sequence model and was thus not compatible with few-shot prompting. We evaluated it by inputting gendered sentences directly and used it as a baseline in all generation experiments.}
\label{tab:models-summary}
\end{table*}

\subsection{Evaluation}
\label{sec:settings-eval}

\paragraph{Test data}
Following \citet{frenda-etal-2024-gfg}, we conduct our GNR experiments on \textsc{mGeNTE} \cite{savoldi2025mgente}, a benchmark for gender-neutral translation from English into several grammatical-gender languages, including Italian. \textsc{mGeNTE} provides 1,500 parallel gendered and gender-neutral references created by professionals
(REF-G and REF-N respectively), differing only in gender expression (see Table \ref{tab:mgente-examples} for an example of an Italian \textsc{mGeNTE} entry).
It is organized into two subsets: \textsc{Set-G}, containing sentences that require neutralization, and \textsc{Set-N}, containing sentences that do not. For our GNR experiments, we use the 750 Italian gendered references from \textsc{Set-N} as input.
Such sentences are ideal input for our task, as they include unnecessary gender specifications by design.

\paragraph{Metrics}
To evaluate gender-neutrality, we use the LLM-as-a-Judge \cite{li2025llm-as-a-judge} approach proposed by \citet{piergentili2025llmasajudge}, which provides sentence-level binary gendered/neutral assessments, and was
shown to be highly accurate in both human- and model-generated texts. We use their optimal configuration for monolingual evaluation.\footnote{Prompt: `\texttt{Mono+P+L}'; GPT model: \texttt{gpt-4o-2024-08-06}} We compute the percentage of neutralized sentences over the whole test set (750 entries).

To evaluate meaning preservation in GNR, we use BERTScore \cite{bert-score}, an attested BERT-based \cite{devlin-etal-2019-bert} 
metric measuring the semantic similarity of two texts (the higher the better, indicating close similarity).
We use BERTScore rather than common string-matching metrics like BLEU \cite{papineni-etal-2002-bleu} and TER \cite{snover-etal-2006-study} because gender-neutralization can have a notable impact on the lexicon, morphology, and structure of a sentence \cite{piergentili-etal-2023-gender}, which would be penalized by such metrics. By contrast, BERTScore was found to be rather insensitive to gender-neutralization \cite{piergentili-etal-2023-hi}. Therefore, lower BERTScore values should be attributed to differences in the meaning of the sentences beyond gender, which we evaluate separately, as described above. 
To identify reference values 
to guide
the interpretation of BERTScore in GNR, we compute the distribution of BERTScore of \textsc{mGeNTE} REF-N sentences against the respective REF-G.\footnote{We only use \textsc{Set-N} entries in this computation.
}
As these neutral reformulations were produced by human experts, the BERTScore distribution provides an empirical estimate of human-level performance in meaning preservation in GNR.
We take the mean BERTScore minus one standard deviation ($0.9334 - 0.0546 = 0.879$) as a reference threshold 
to obtain a conservative estimate of human-level performance. This accounts for the natural variability in human reformulations while filtering out outliers with lower semantic similarity, thus ensuring that models evaluated above this threshold perform within the typical human range.

\begin{table*}
    \centering
    \footnotesize
    \begin{tabularx}{\textwidth}{>{\raggedright\arraybackslash}X>{\raggedright\arraybackslash}X}
    \toprule
       \textbf{\textsc{GFG}, Italian}  & \textbf{\textsc{GFG}, English} \\
    \midrule
        Riformula la seguente frase utilizzando un linguaggio neutro rispetto al genere dei referenti umani, evitando l'uso di forme maschili e femminili. & Rewrite the following Italian sentence using a gender-neutral language in reference to human beings, avoiding masculine or feminine forms.\\
    \midrule
    \textbf{\textsc{Rewrite}, Italian}  & \textbf{\textsc{Rewrite}, English} \\
    \midrule
    Sei un riscrittore di frasi italiane con l'obiettivo di rendere i testi neutrali rispetto al genere dei referenti umani.
Ti viene fornita una frase che contiene riferimenti a persone in forme marcate per genere, come il maschile sovraesteso o coppie binarie.

Il tuo compito è riformulare la frase in modo da:
\begin{itemize}
\item rimuovere riferimenti espliciti al genere quando non necessari;
\item mantenere inalterato il significato originale;
\item preservare lo stile e la leggibilità del testo.
\end{itemize}

Per farlo, usa strategie come:
\begin{itemize}
\item sostantivi collettivi (``la cittadinanza'', ``il personale'', ``l'utenza'');
\item perifrasi impersonali (``si dovrebbe'', ``si consiglia'');
\item forme passive (``l'accesso è consentito'');
\item forme imperative (``allega il documento'');
\item pronomi relativi e costruzioni subordinate (``chi ha svolto attività di pesca'');
\item termini epiceni (``ogni giudice'', ``gentile collega'');
\item termini neutri (``l'individuo'', ``la persona interessata'', ``il membro'').
\end{itemize}

IMPORTANTE:
\begin{itemize}
\item evita l'uso del maschile come forma generica e non usare forme grafiche non standard come asterischi o schwa;
\item evita doppie formulazioni come ``il/a cittadino/a'' oppure ``il professore o la professoressa'';
\item non rimuovere parti della frase che non richiedono modifiche (ad esempio, i nomi propri);
\item fornisci solo la frase riformulata.
\end{itemize}
& 
You are a rewriter of Italian sentences with the goal of making texts gender-neutral with respect to human referents.
You are given a sentence that contains references to people using gender-marked forms (such as masculine generics or binary pairs).

Your task is to rewrite the sentence to:
\begin{itemize}
\item remove explicit gender references when they are not necessary;
\item preserve the original meaning;
\item maintain the style and readability of the text.
\end{itemize}

To do this, use strategies such as:
\begin{itemize}
\item collective nouns (``la cittadinanza'', ``il personale'', ``l'utenza'');
\item impersonal phrases (``si dovrebbe'', ``si consiglia'');
\item passive constructions (``l'accesso è consentito'');
\item imperative constructions (``allega il documento'');
\item relative pronouns and subordinate clauses (``chi ha svolto attività di pesca'');
\item epicene terms (``ogni giudice'', ``gentile collega'');
\item neutral terms (``l'individuo'', ``la persona interessata'', ``il membro'').
\end{itemize}

IMPORTANT:
\begin{itemize}
\item avoid using the masculine form as a generic and do not use non-standard spellings such as asterisks or schwa;
\item avoid binary formulations such as ``il/a cittadino/a'' or ``il professore o la professoressa'';
\item do not remove any part of the sentence that does not need to be rewritten (e.g. proper names);
\item only return the reformulated sentence.
\end{itemize} \\
    \bottomrule
    \end{tabularx}
    \caption{The `system' role messages for the two prompt formats used in the few-shot prompting experiments, in both Italian and English.}
    \label{tab:prompts}
\end{table*}

\subsection{Models}
\label{sec:settings-models}

We experiment with a diverse set of models spanning different families, architectures, scales, and language coverage. Table \ref{tab:models-summary} summarizes our selection of models and how we use them in our experiments. Our selection includes:

\begin{itemize}
    \item \textbf{`Italian' models}, specifically designed or adapted for Italian language tasks: Minerva 7B, LLaMAntino 8B, and Velvet 14B.
    \item \textbf{Multilingual LLMs}, trained on multiple languages including Italian, to evaluate 
    general-purpose models: Llama 3.1 8B, Phi 4 14B, Llama 3.3 70B. Among the multilingual LLMs, we include four different-sized models from the \textbf{Qwen3 family}, 
    to analyze consistency and scalability within a single architecture.
    \item One \textbf{commercial system}, included as a high-performance reference system: GPT-4.1.\footnote{Model \texttt{gpt-4.1-2025-04-14}}
    \item Inclusively,\footnote{\url{https://huggingface.co/E-MIMIC/inclusively-reformulation-it5}} a fine-tuning of \texttt{it5-large} \cite{sarti-nissim-2024-it5-text}, as the only \textbf{dedicated model} for Italian GNR. We consider this system the baseline for our experiments.
\end{itemize}

All models, except for Inclusively, are instruction-tuned autoregressive LLMs. 

\subsection{Few-Shot Prompting}
\label{sec:settings-prompting}

We run few-shot prompting experiments with all models in the selection described above,\footnote{Except for Inclusively, which does not support few-shot prompting. We instead test its off-the-shelf generation capabilities.} to investigate the performance of LLMs without any task-specific fine-tuning. We use two prompt formats:

\begin{itemize}
    \item \textbf{\textsc{GFG}}: a concise rewriting instruction, originally used by \citet{frenda-etal-2024-gfg} in their gender-fair generation challenge for Italian LLMs.
    \item \textbf{\textsc{Rewrite}}: a more detailed and analytical prompt, also featuring essential guidelines for the task with neutralization examples following the strategies identified by \citet{piergentili-etal-2023-gender}.
\end{itemize}

These prompts allow us to explore the impact of more complex instruction on models' performance.
Moreover, we experiment with these two prompt formats by formulating them in both Italian an English, to investigate whether the language used 
is a relevant factor as well. The content of the prompts is reported in Table \ref{tab:prompts}.
We include 
the same 8 task exemplars---or 
shots---with all prompts, to elicit the in-context learning ability of LLMs \cite{brown-2020-few-shot}. We use \texttt{vLLM} \cite{kwon2023vllm} as the inference engine. 

\subsection{Fine-tuning}
\label{sec:settings-finetuning}

\begin{figure}
    \centering
    \includegraphics[width=1\linewidth]{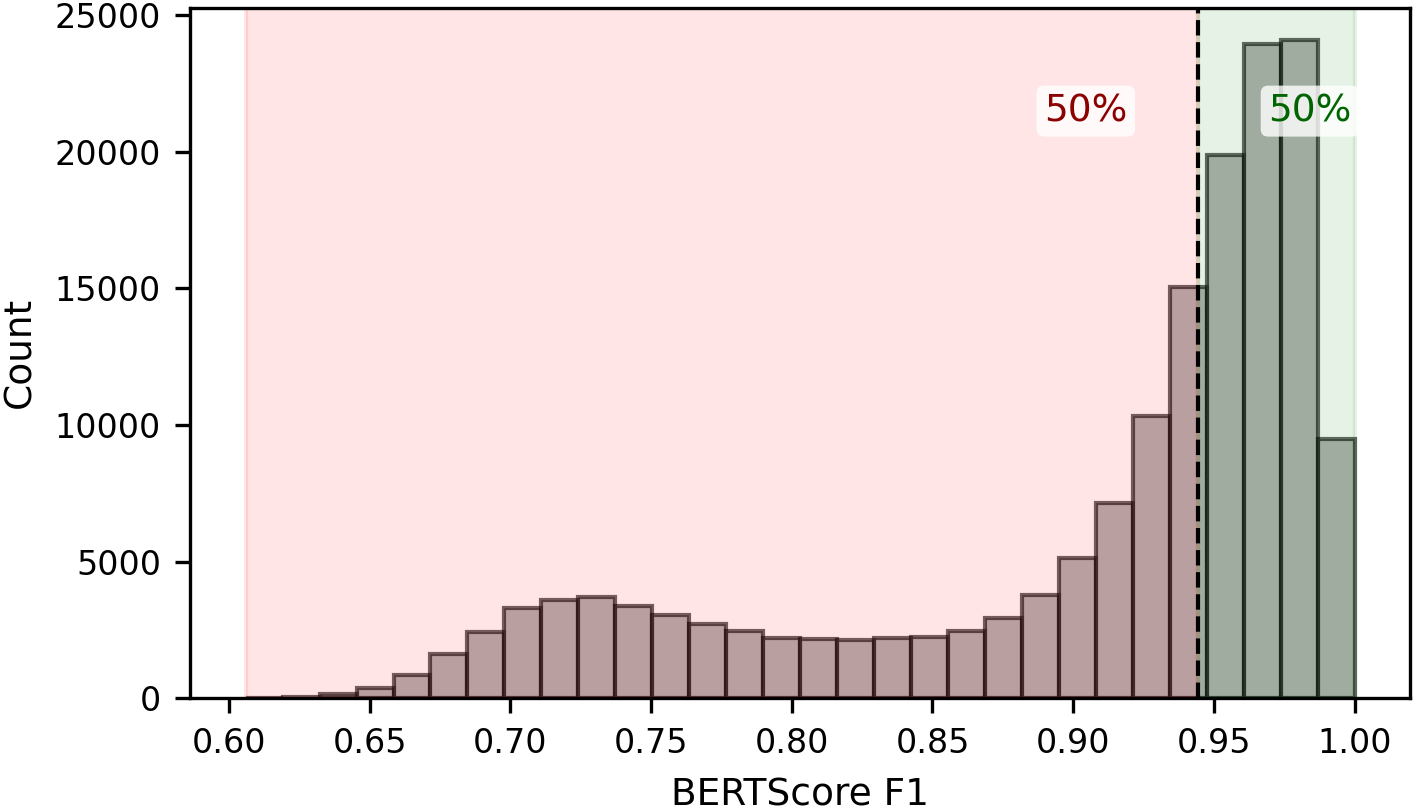}
    \caption{Distribution of BERTScore values over the \textsc{\color{RoyalBlue}{full}} fine-tuning dataset. 
    The \textsc{\color{ForestGreen}{clean}} split is also visualized as the green portion starting at the median line (0.9443).
    }
    \label{fig:data-selection}
\end{figure}

\begin{table}
\footnotesize
    \centering
    \begin{tabular}{cccc}
        \textbf{Set} & \textbf{Entries} & \textbf{Selection criterion} & \textbf{Avg. BERTScore} \\
        \toprule
       {\textsc{\color{RoyalBlue}{full}}}  & 162,778  & - & 0.9044 \\
       {\textsc{\color{ForestGreen}{clean}}} & 81,389 & BERTScore $\geq$ median & 0.9697 \\
       \bottomrule
    \end{tabular}
    \caption{Training datasets statistics and summary.}
    \label{tab:ft-data-stats}
\end{table}

\begin{figure}
    \centering
    \includegraphics[width=1\linewidth]{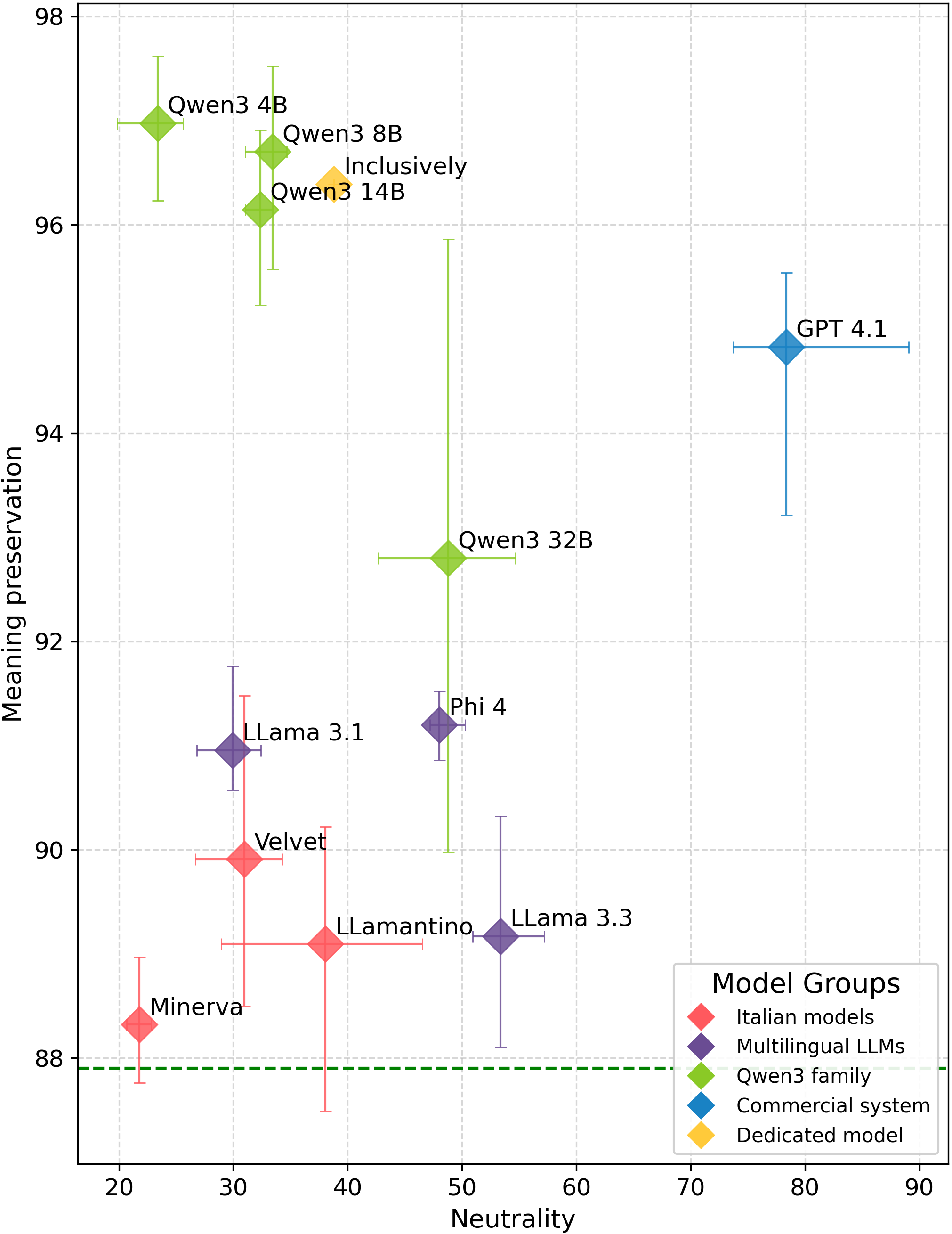}
    \caption{\textbf{Results of the few-shot prompting experiments.} 
    The meaning preservation (vertical) axes report BERTScore values multiplied by 100 for easier visualization, whereas the neutrality (horizontal) axes report sentence-level neutralization accuracy.
    Each $\Diamond$ represents the average performance of a model across four prompts.
    The lines extending from each $\Diamond$  indicate the full range of values observed for that model on the respective axis. The dashed line 
    indicates the reference value for human-level 
    meaning preservation in GNR.}
    \label{fig:prompting-results}
\end{figure}

We perform fine-tuning experiments to assess whether and to which extent smaller open-weight LLMs can be adapted to the GNR task and approach the performance of larger models or closed systems.
Namely, we fine-tune LLaMAntino, Velvet, LLama 3.1, Phi 4, and the 8B and 14B Qwen3 models.

\subsubsection{Data}
\label{sec:finetuning-data}
The only openly available development data dedicated to Italian GNR is the synthetically generated training dataset used 
by \citet{piergentili-etal-2023-hi} to train a gender-neutrality classifier.\footnote{More specifically, we use the cleaned version of the dataset later released by \citet{savoldi-etal-2024-prompt} at \url{https://github.com/hlt-mt/fbk-NEUTR-evAL/blob/main/solutions/GeNTE.md}}
This data consists in gendered Italian sentences and their gender-neutral counterparts, all generated starting from a dictionary of masculine, feminine, and neutral expressions, through a multi-step prompting pipeline. 
We repurpose this data to fine-tune autoregressive LLMs for GNR. We prepare the data as chat-formatted input, where each instance consists of a \textit{user} role message containing a gendered sentence, and an \textit{assistant} role message containing the corresponding neutral sentence. Consistent with the models’ prior instruction‐following fine‐tuning, this method adopts a conversational prompt–response format while strictly adhering to a causal token‐prediction objective \cite{ouyang2022instructgpt}.

As the sentences were partly LLM-generated, 
we note that the content of the gendered-neutral pairs may not always be aligned due to the unpredictability of LLMs in open-ended generation.\footnote{While this is not necessarily an issue in the development of a classifier, where individual sentences are simply paired with neutrality labels, for a rewriting task the input-output sentences should be identical except for the attribute of interest, i.e., in this case, gender.}
To investigate this aspect, we compare the gendered and neutral sentences in the dataset using BERTScore 
to identify dataset entries with semantically divergent gendered-neutral sentences.
Figure \ref{fig:data-selection} reports the BERTScore values for the entire dataset. We observe that while the score distribution is skewed towards almost-perfect values, there is a notable tail of 
gendered-neutral
sentence pairs with a rather divergent semantic content. To investigate the impact of such data in GNR fine-tuning, 
we construct a subset to be used for training alongside the {\textsc{\color{RoyalBlue}{full}}} dataset: a {\textsc{\color{ForestGreen}{clean}}} subset obtained by filtering out the bottom 50\% of sentence pairs based on the BERTScore values.
Statistics about the fine-tuning data are reported in Table \ref{tab:ft-data-stats}.

\subsubsection{Method}
\label{sec:method}
We fine-tune the selected models 
using Low-Rank Adaptation (LoRA) \cite{hu2021lora}.
Following common practices in LoRA fine-tuning \ap{\cite{unsloth-lora-hyperparams-guide-2025}} we set the \texttt{rank} and \texttt{alpha} at 32, and use the following hyperparameters to strike a balance between hardware constraints\footnote{We run our experiments on nodes with 4 NVIDIA A100 GPUs with 64 GB VRAM each.} and consistency across model sizes and requirements: \texttt{learning rate}: $2 \times 10^{-4}$, \texttt{batch size}: 8 for the 8B models, 4 for the 14B models.
We use early stopping with a patience of 20 steps for the 8B models and 40 steps for the 14B models.

\section{Results}
\label{sec:results}

\begin{figure*}
    \centering
    \includegraphics[width=1\linewidth]{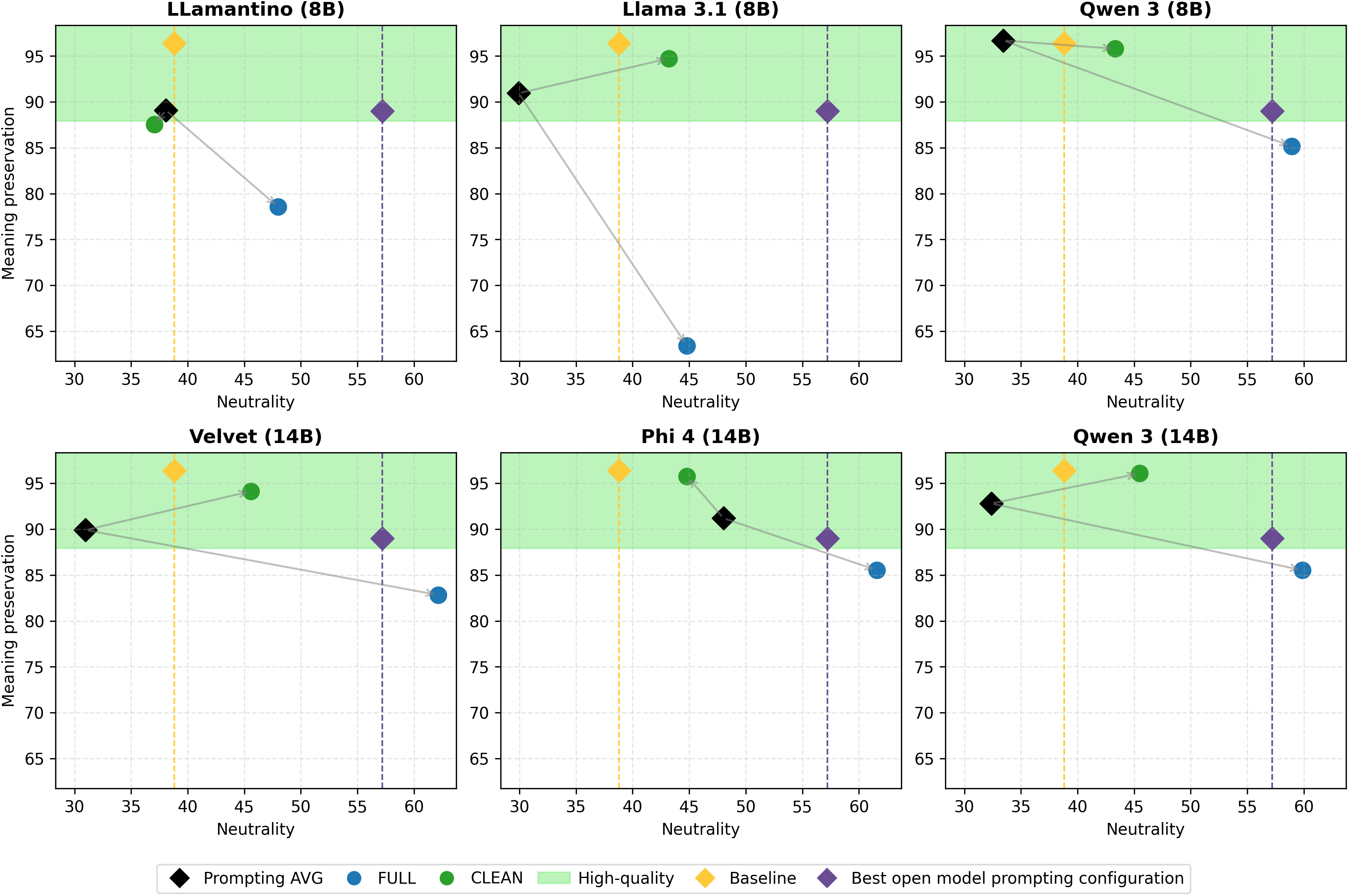}
    \caption{\textbf{Results of the fine-tuning experiments.} The meaning preservation (vertical) axes report BERTScore values multiplied by 100 for easier visualization, whereas the neutrality (horizontal) axes report sentence-level neutralization accuracy. The black diamond represents the average performance of the model in the prompting experiments. The blue and green points represent the performance of the model fine-tuned on the {\textsc{\color{RoyalBlue}{full}}} and {\textsc{\color{ForestGreen}{clean}}} datasets respectively. The green band at the top represents BERTScore values reaching human-level meaning preservation in GNR. The yellow and blue points and dashed vertical lines respectively represent the baseline (the dedicated model Inclusively) and the best configuration performance of an 
    open-weight model (LLama 3.3 70B, GFG English prompt).}
    \label{fig:finetuning-results}
\end{figure*}

\subsection{Few-Shot Prompting Results}
\label{sec:prompting-results}

Figure \ref{fig:prompting-results} summarizes the results of the few-shot prompting experiments showing all models’ performance in neutrality and meaning preservation. Higher values on both axes indicate better 
performance; therefore, systems closer to the top-right corner perform best. 
As no consistent trend emerged across prompt formats (\textsc{GFG} vs. \textsc{Rewrite}, see Section \ref{sec:settings-prompting}) and languages (Italian vs. English),
we report each model’s average performance, along with the range of neutrality and BERTScore values observed across prompting conditions. 
In Appendix \ref{app:detailed-results-prompting} we provide the complete and detailed results obtained with the two prompt formats, separately for Italian and English instructions.

Generally, and with rare exceptions, all models' BERTScore values are well above the quality threshold we identified in §\ref{sec:settings-eval}. This means that the models do not generate unrelated or additional text, confirming that their outputs remain adherent to the input and free of ``hallucinations'' \cite{Huang_2025}.

Neutrality scores, 
on the contrary, vary significantly across models. Looking at our baseline, the GNR-dedicated model Inclusively, we observe that it performs rather poorly in neutrality. 
Across LLMs, we notice similar behavior within the groups. The ``Italian'' models, in the bottom left quarter of the chart, generally fail to neutralize, and alter the sentences the most. Within the multilingual LLMs group, only Phi 4, Qwen3 32B, and LLama 3.3 perform better than the Italian models. The rest of the Qwen3 family generally underperforms, with the high BERTScore suggesting that they make little to no change to the gendered sentences. The only model performing well on both axes is GPT 4.1, which tops at 89.07\% neutralization accuracy and 93.21 BERTScore, indicating that it correctly alters the parts of the sentences expressing the gender of human beings while leaving the rest untouched.

Overall, we find that the LLMs we tested perform very differently in GNR in Italian, and that failure in this setting consists in overlooking the relevant (gendered) parts of the input to act upon, and/or unsuccessfully rendering them gender-neutral.

\subsection{Fine-Tuning Results}
\label{sec:fine-tuning-results}

\begin{figure*}
    \centering
    \includegraphics[width=1\linewidth]{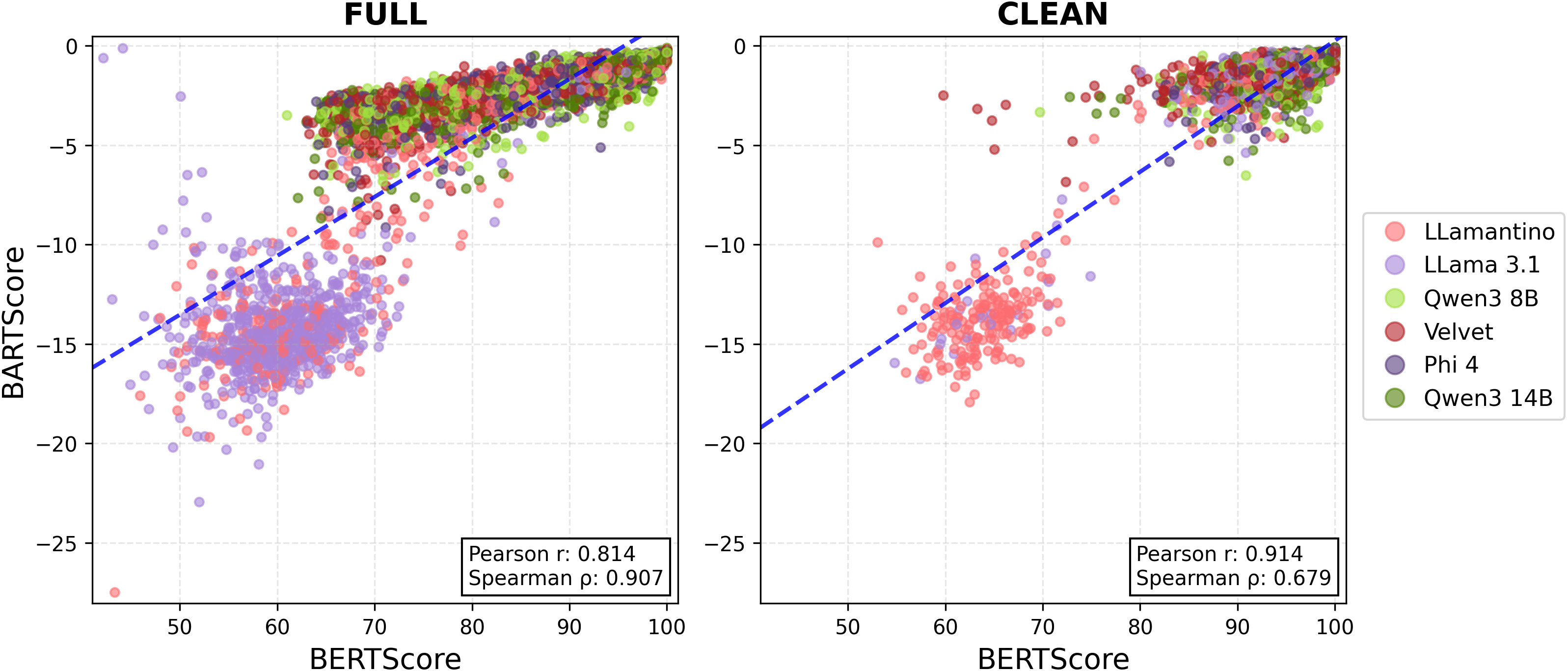}
    \caption{\textbf{BERTScore and BARTScore for the outputs of the models fine-tuned on both} {\textsc{\color{RoyalBlue}{full}}} \textbf{and} {\textsc{\color{ForestGreen}{clean}}}. The dashed lines are least-squares regression lines fitted to each set of points, modeling the relationship between the metrics. Points above the line have higher BARTScore than predicted by BERTScore (i.e. BERTScore underrates them), and vice versa for points below. We report Pearson \textit{r} and Spearman $\rho$ correlation coefficients for each split as well.}
    \label{fig:bert-bart}
\end{figure*}

Results of the fine-tuning experiments are reported in Figure \ref{fig:finetuning-results}. We first notice that on the neutrality axis all fine-tuned models outperform the baseline, except for LLamantino/{\textsc{\color{ForestGreen}{clean}}} configuration. 
LLamantino shows the narrowest gains overall, and in one case even a drop in neutrality, echoing its weaker few-shot prompting results and suggesting it may be ill-suited to GNR.
In four out of six instances, and always with the {\textsc{\color{RoyalBlue}{full}}} dataset, the fine-tuned models also outperform the best performer among the 
open-weight models in the prompting experiments, i.e. LLama 3.3 70B with the GFG English prompt, though with a significant drop in BERTScore. 

Such a drop indicates that these models fail by hallucinating unrelated content in their attempt to neutralize, rather than by leaving the input sentences untouched as observed in the prompting experiments (§\ref{sec:prompting-results}). This is possibly due to two factors: the significantly smaller size of the fine-tuned models with respect to LLama 3.3 70B (1/9 or 1/5, for the 8B and 14B models respectively), as larger LLMs have been shown to exhibit greater robustness and lower variance in downstream performance after fine-tuning compared to smaller counterparts \cite{chung2024scaling}, and/or the presence of many divergent gender-neutral sentence pairs in the fine-tuning dataset (see §\ref{sec:finetuning-data}).

While {\textsc{\color{RoyalBlue}{full}}} yields the highest improvements in neutrality, only {\textsc{\color{ForestGreen}{clean}}} improves performance on both axes while keeping BERTScore within the human-level range. However, it yields significantly smaller gains in neutrality and even causes drops for two models (LLamantino, Phi 4). We hypothesize that {\textsc{\color{ForestGreen}{clean}}} may be excessively conditioned by the data filtering method, i.e. a BERTScore based selection.
In other words, by selecting only dataset entries with almost perfect BERTScore values we are optimizing the models to perform well on the sentence similarity dimension---as measured by BERTScore---rather than GNR.

\paragraph{The impact of metric-based data selection}
To investigate the hypothesis above, we evaluate the same outputs against the gendered inputs with another semantic similarity metric: BARTScore \cite{yuan2021bartscore}.\footnote{While similar in name and scope, BERTScore and BARTScore function differently. The first computes a sum of token-level cosine similarities between two sentences' embeddings encoded by a BERT (encoder-only) model; the latter is computed as the weighted sum of the log‐probabilities that a pretrained BART (encoder-decoder) model assigns to each token in the generated text. In our experiments, we use the BART model \href{https://huggingface.co/facebook/bart-large}{\texttt{facebook/bart-large}} \cite{lewis2019bart}.}
BERTScore and BARTScore evaluations are visualized in Figure \ref{fig:bert-bart}.
To understand whether outputs of the models fine-tuned on {\textsc{\color{ForestGreen}{clean}}} are actually very semantically similar to the corresponding input, and whether those models simply learned to game BERTScore, we compute\footnote{We use the Python library \texttt{SciPy} \cite{2020SciPy-NMeth}.} the Pearson \textit{r} and Spearman $\rho$ correlation coefficients between BERTScore and BARTScore assessments.
The first captures linear correlations between the two metrics’ raw scores, while the latter measures how well the relationship between the two variables can be described by a monotonic function, by comparing the rankings of the scores rather than their raw values. This combination allows us to assess both the alignment of the scores and the consistency in how the two metrics rank the outputs.

We find that in {\textsc{\color{RoyalBlue}{full}}}, \textit{r} equals 0.814 and $\rho$ equals 0.907, whereas in {\textsc{\color{ForestGreen}{clean}}} they are 0.914 and 0.679 respectively.\footnote{All p-values $< 0.05$.} \textit{r} is high in both cases, indicating a strong linear correlation between the two metrics---stronger in {\textsc{\color{ForestGreen}{clean}}}, as in that case the data points are more tightly clustered, skewed towards higher values. This confirms that the metrics generally agree on the quality of the outputs. 
The substantial drop in $\rho$, instead, indicates that there are many instances in {\textsc{\color{ForestGreen}{clean}}} where the monotonic trend is broken, i.e., higher BERTScore does not necessarily correspond to higher BARTScore. This suggests that the {\textsc{\color{ForestGreen}{clean}}} models also learned to game BERTScore by reproducing features rewarded by that metric. 

With respect to our hypothesis: by selecting high-similarity pairs for the {\textsc{\color{ForestGreen}{clean}}} dataset, we effectively steered models toward preserving semantic alignment with the input; however, this emphasis on similarity appears to have hampered their improvement in neutralization. Indeed, the models learned to preserve the input to an excessive degree, as confirmed by the high \textit{r} coefficient and high BARTScore values shown in Figure \ref{fig:bert-bart}. We interpret our results as evidence of a broader trade-off between optimizing for neutrality and for sentence similarity.  
Our findings underscore the need for data curation strategies that strike a balance between neutrality and similarity, achieving the flexibility required for effective GNR.

\section{Conclusions}

We presented the first systematic investigation of state-of-the-art large language models for Italian gender-neutral rewriting under a two-dimensional evaluation of neutrality and meaning preservation. In our few-shot prompting experiments, 
open-weight
models outperformed the only existing Italian-specific system but remained behind a closed commercial system.  

Through fine-tuning experiments 
we showed that compact models can match or exceed the best 
open-weight
LLM at a fraction of its size. Moreover, our BERTScore-based data cleaning highlighted a trade-off: models  trained on cleaned data achieve human-level BERTScore but show smaller neutrality gains and exhibit ranking differences against another similarity metric, signaling over-fitting on BERTScore. 
Future work should take this trade-off into account and create dedicated, high-quality parallel data to aim at reaching the performance of the commercial system with  
open-weight
models.

\begin{acknowledgments}
We acknowledge the support of the project  InnovAction: Network Italiano dei Centri per l’Innovazione Tecnologica (CUP B47H2200437000), funded by MIMIT with NPRR - NextGenerationEU funds, in collaboration with Piazza Copernico S.r.l. 
We also received funding from the PNRR project FAIR - Future AI Research (PE00000013), under the NRRP MUR program funded by NextGenerationEU.
Finally, we acknowledge the CINECA award under the ISCRA initiative (AGeNTE), for the availability of high-performance computing resources and support.
\end{acknowledgments}


\bibliography{sample-ceur,custom}

\appendix
\section{Detailed results}
\label{app:detailed-results-prompting}

Tables \ref{tab:prompting-neutrality} and \ref{tab:prompting-bertscore} report the detailed results of our fine-tuning experiments.

\begin{table*}[ht]
\centering
\begin{tabular}{lccccccc}
\multicolumn{8}{c}{\textbf{\textsc{Neutrality}}} \\
\toprule
\multicolumn{2}{c}{\textbf{Model}} 
  & \textbf{Size (B)} 
  & \textbf{GFG Ita} 
  & \textbf{GFG Eng} 
  & \textbf{Rewrite Ita} 
  & \textbf{Rewrite Eng} 
  & \textbf{AVG} \\
\midrule

\multirow{3}{*}{\textbf{"Italian" models}} 
  & Minerva     &  7  & 20.67 & \underline{22.80} & 22.67 & 21.07 & 21.80 \\
  & LLaMAntino  &  8  & 28.93 & 31.07 & \cellcolor{TealBlue!30} \underline{46.53} & 45.73 & 38.07 \\
  & Velvet      & 14  & 32.40 & \underline{34.27} & 30.53 & 26.67 & 30.97 \\
\midrule

\multirow{3}{*}{\textbf{Multilingual LLMs}} 
  & Llama 3.1   &  8  & 26.80 & 28.27 & 32.27 & \underline{32.40} & 29.93 \\
  & Phi 4       & 14  & 47.47 & 47.20 & 47.20 & \underline{50.27} & 48.03 \\
  & Llama 3.3   & 70  & 52.93 & \cellcolor{TealBlue!30} \underline{57.20} & 52.40 & 50.93 & 53.37 \\
\hdashline

\multirow{4}{*}{\textbf{Qwen3 family}} 
  & Qwen3   &  4  & 23.87 & 19.87 & \underline{25.60} & 24.27 & 23.40 \\
  & Qwen3  &  8  & 33.60 & \underline{34.67} & 34.40 & 31.07 & 33.43 \\
  & Qwen3  & 14  & 32.27 & 31.07 & \underline{33.47} & 32.67 & 32.37 \\
  & Qwen3  & 32  & \cellcolor{TealBlue!40} \underline{54.67} & 52.80 & 42.67 & 45.07 & 48.80 \\
\midrule

\multirow{1}{*}{\textbf{Commercial system}} 
  & GPT 4.1   
    & \multicolumn{1}{c}{?}    
    & 75.33 & \textbf{\underline{89.07}} & 73.73 & 75.33 & 78.37 \\
\midrule

\multirow{1}{*}{\textbf{Dedicated model}} 
  & Inclusively 
    &  0.78  
    & \multicolumn{4}{c}{38.80}
    & 38.80 \\
\bottomrule
\end{tabular}
\caption{\textbf{Neutrality results of the few-shot prompting experiments.} The best model settings are \underline{underlined}, the best settings across the categories are \colorbox{TealBlue!40}{highlighted}, and the best overall performer is in \textbf{bold}.}
\label{tab:prompting-neutrality}
\end{table*}

\begin{table*}[ht]
\centering
\begin{tabular}{lccccccc}
\multicolumn{8}{c}{\textbf{\textsc{BERTScore}}} \\
\toprule
\multicolumn{2}{c}{\textbf{Model}} 
  & \textbf{Size (B)} 
  & \textbf{GFG Ita} 
  & \textbf{GFG Eng} 
  & \textbf{Rewrite Ita} 
  & \textbf{Rewrite Eng} 
  & \textbf{AVG} \\
\midrule

\multirow{3}{*}{\textbf{"Italian" models}} 
  & Minerva     &  7  & 87.78 & 88.78 & 87.76 & \underline{88.97} & 88.32 \\
  & LLaMAntino  &  8  & 89.97 & \underline{90.22} & 87.49 & 88.70 & 89.09 \\
  & Velvet      & 14  & 89.60 & \cellcolor{TealBlue!40} \underline{91.48} & 88.50 & 90.06 & 89.91 \\
\midrule

\multirow{3}{*}{\textbf{Multilingual LLMs}} 
  & Llama 3.1   &  8  & \cellcolor{TealBlue!40} \underline{91.76} & 90.70 & 90.78 & 90.57 & 90.95 \\
  & Phi 4       & 14  & 90.86 & 90.95 & \underline{91.52} & 91.46 & 91.20 \\
  & Llama 3.3   & 70  & 88.10 & 89.00 & 89.26 & \underline{90.32} & 89.17 \\
\hdashline

\multirow{4}{*}{\textbf{Qwen3 family}} 
  & Qwen3 &  4  & 96.23 & 96.98 & 97.07 & \cellcolor{TealBlue!40} \textbf{\underline{97.62}} & 96.97 \\
  & Qwen3 &  8  & 96.49 & 95.57 & 97.23 & \underline{97.52} & 96.70 \\
  & Qwen3 & 14  & 95.23 & 96.72 & 95.72 & \underline{96.91} & 96.14 \\
  & Qwen3 & 32  & 89.98 & 91.31 & 94.04 & \underline{95.86} & 92.80 \\
\midrule

\multirow{1}{*}{\textbf{Commercial system}} 
  & GPT 4.1   
    & \multicolumn{1}{c}{?}   
    & 95.12 
    & 93.21 
    & \underline{95.54}
    & 95.44 
    & 94.83 \\
\midrule

\multirow{1}{*}{\textbf{Dedicated model}} 
  & Inclusively 
    &  0.78  
    & \multicolumn{4}{c}{96.39}
    & 96.39 \\
\bottomrule
\end{tabular}
\caption{Sentence-similarity results of the few-shot prompting experiments. The best model settings are \underline{underlined}, the best settings across the categories are \colorbox{TealBlue!40}{highlighted}, and the best overall performer is in \textbf{bold}.}
\label{tab:prompting-bertscore}
\end{table*}

\end{document}